\documentclass{article}

\usepackage{microtype}
\usepackage{graphicx}
\usepackage{booktabs}
\usepackage{hyperref}

\usepackage[accepted]{icml2026}

\usepackage{amsmath}
\usepackage{amssymb}
\usepackage{mathtools}
\usepackage{amsthm}
\usepackage{enumitem}

\setlist[itemize]{leftmargin=*,topsep=2pt,itemsep=1pt}
\setlist[enumerate]{leftmargin=*,topsep=2pt,itemsep=1pt}

\icmltitlerunning{Can Training Logs Make Model Comparisons More Precise?}

\begin{document}

\twocolumn[
  \icmltitle{Can Training Logs Make Model Comparisons More Precise?}

  \begin{icmlauthorlist}
    \icmlauthor{Wei-Jung Huang}{ind}
  \end{icmlauthorlist}

  \icmlaffiliation{ind}{Independent Researcher}

  \icmlcorrespondingauthor{Wei-Jung Huang}{william.wj.huang@gmail.com}

  \icmlkeywords{variance reduction, covariate adjustment, hypothesis testing, model comparison, stochastic training}

  \vskip 0.3in
]

\printAffiliationsAndNotice{}

\begin{abstract}
Comparing stochastically trained models requires estimating both a performance
difference and its uncertainty from repeated runs. We study whether training
logs from those same runs can make such comparisons more precise. Because
training-log covariates are produced during training rather than measured
before it, we use arm-specific covariate adjustment: each model is adjusted
only with statistics from its own runs, and the raw mean difference remains the
reported effect. In a vision study spanning three architectures and three
datasets, simple adjustments based on early training logs often reduce
uncertainty in model comparisons. The main limitation is covariate selection.
Broadly searching the log pool for the most correlated statistic often adds
more noise than it removes, even when useful statistics exist in hindsight.
Training logs therefore appear useful for more precise model comparisons, but
only when the adjustment avoids large selection noise.
\end{abstract}

\section{Introduction}
\label{sec:intro}

Every comparison between stochastically trained models rests on repeated runs.
Because each run depends on initialization, data order, and nondeterministic
implementation details, the performance difference between two models is not
a single number but a statistical estimate with uncertainty. Run-to-run
variation directly determines the standard error of this estimate and can
change reported confidence intervals and model
rankings~\citep{bouthillier2021accounting,henderson2018deep,dehghani2021benchmark}.
The standard remedy is to run more repetitions, but this scales linearly in
compute.

If early training statistics correlate with final test accuracy across
repeated runs, subtracting their centered contribution can reduce run-level
variation. Applied separately to each model, this adjustment can tighten the
confidence interval for the performance difference without changing which
models are being compared. The idea is closely related to regression
adjustment (CUPED) in online A/B testing~\citep{deng2013cuped}, where
pre-experiment user metrics reduce outcome variance and thereby tighten the
confidence interval for the treatment effect.

An important difference is that A/B testing covariates are measured before
treatment assignment, whereas training-log covariates such as early validation
accuracy, gradient norms, and batch-loss summaries are co-produced with the
outcome. Because these covariates may differ across models for scientific
reasons, a pooled adjustment could remove part of the model difference itself.
We therefore use \emph{arm-specific} adjustment: each model is
adjusted using only its own training-log covariates, and the adjusted
dispersions are combined in the standard error for the performance difference.

We address three empirical questions: whether training logs contain
useful run-level signal about final performance, whether this signal can be
selected reliably at typical run budgets, and whether arm-specific
adjustment tightens confidence intervals for model-performance differences
when the statistic used for adjustment is fixed in advance. The experimental design is a
$3\times 3$ factorial study (ResNet-18, ViT-Tiny, ConvNeXt-Tiny $\times$
CIFAR-10, CIFAR-100, Tiny-ImageNet; 50 runs per cell, 450 runs total). Our
contributions are:
\begin{enumerate}
    \item \textbf{An arm-specific adjustment framework for model comparison.}
    Each model is adjusted using only its own training-log covariates, with
    cross-fitted coefficient estimation, coverage diagnostics, and null
    calibration checks
    (\S\ref{sec:method}, \S\ref{sec:coverage}, \S\ref{sec:null}).
    \item \textbf{Model-specific covariate signal and selection risk.}
    A fixed early validation-loss adjustment and PCA summaries of training-log
    families can reduce single-arm variance, in some cells by sizable margins,
    but which training statistic helps depends on the model and training recipe, and
    selecting the best one from a large candidate pool often backfires
    (\S\ref{sec:vr}, \S\ref{sec:selection_failure}).
    \item \textbf{Pairwise comparison evidence.} Arm-specific adjustment
    narrows confidence intervals for performance differences at sufficient
    run budgets but widens them when runs are too few for stable estimation
    (\S\ref{sec:pairwise}, \S\ref{sec:budget}).
    \item \textbf{Practical guidance.} A training-log statistic should be
    chosen before it is used as the primary adjustment for a model comparison;
    otherwise, the adjustment should be reported as exploratory. Reliable
    selection under limited run budgets remains the main open problem
    (\S\ref{sec:discussion}).
\end{enumerate}

\section{Related Work}
\label{sec:related}

\paragraph{Run-to-run variation in deep learning.}
The sensitivity of deep learning results to random initialization has been
documented in reinforcement learning~\citep{henderson2018deep},
GANs~\citep{lucic2018gans}, NLP fine-tuning~\citep{dodge2020fine,sellam2022multiberts},
and supervised vision~\citep{bouthillier2021accounting,picard2021torch}.
\citet{dehghani2021benchmark} argued that run-to-run variation and single-split
evaluation create a ``benchmark lottery'' in which perceived rankings are
fragile. These studies motivate running more repetitions or reporting variance
more carefully. Complementing this line of work, we ask whether training logs
can reduce uncertainty from a fixed run budget.

\paragraph{Regression adjustment and variance reduction.}
Classical variance-reduction methods subtract a correlated signal with known
mean to reduce Monte Carlo variance~\citep{owen2013monte}.
Regression adjustment~\citep{deng2013cuped} applies this in A/B testing using
pre-experiment user metrics, with nonlinear extensions via boosted
trees~\citep{poyarkov2016boosted}. In the randomized-experiment literature,
\citet{freedman2008regression} showed that regression adjustment can improve
or worsen precision, can introduce finite-sample bias, and can make usual
standard errors misleading under randomization.
\citet{lin2013agnostic} showed that these concerns are minor or fixable in
large samples when the adjustment interacts treatment with centered covariates
and uses robust standard errors. Our setting differs in two respects. First,
the estimand is a model's expected performance over repeated training runs,
and pairwise comparisons combine two such arm-specific estimators. Second, the
covariates are co-produced rather than pre-treatment. Out-of-fold estimation
mitigates direct overfitting, but the co-produced setting lacks the
pre-treatment independence guarantee, so we check interval calibration
by resampling from the 50 runs we actually trained
(Section~\ref{sec:coverage}). Ridge and James-Stein shrinkage
could further stabilize coefficient estimates in the multivariate case; we
focus on ordinary least squares (OLS) and principal component analysis (PCA)
for interpretability.
Bayesian partial pooling~\citep{gelman2013bayesian} offers an alternative; we
adopt the frequentist approach because it requires fewer distributional
assumptions and gives interval estimates whose coverage can be checked
empirically in our diagnostics.

\paragraph{Learning curve prediction.}
Early-training signals predict converged
performance~\citep{domhan2015speeding}, a structure routinely exploited for
early stopping and hyperparameter search. We repurpose the same correlations
for a different goal: rather than predicting final accuracy or deciding when to
stop, we use them to reduce run-level uncertainty after training completes.
The fixed validation-loss covariate in our experiments (validation loss at the
first-third epoch) is a direct application of this learning-curve structure
for covariate adjustment. No modification to the training pipeline is needed.

\section{Method}
\label{sec:method}

\subsection{Arm-Specific Covariate Adjustment}

Let model $m \in \{A,B\}$ produce final test accuracy $Y_{m,s}$ from run
$s \in \{1,\ldots,n_m\}$, and let $X_{m,s} \in \mathbb{R}^{p_m}$ be
training-log covariates from the same run. The model-comparison estimand is
\begin{equation}
    \Delta = \mu_A - \mu_B,\quad
    \mu_m = \mathbb{E}[Y_{m,s}],
\end{equation}
where the expectation is over repeated stochastic training runs for the same
model and recipe. The null hypothesis is $H_0\colon \Delta = 0$. The reported
point estimate of $\Delta$ is always the raw mean difference
$\bar{Y}_A - \bar{Y}_B$; covariate adjustment changes only the estimated
standard error and confidence interval width, not the point estimate itself.

For a fixed training-log statistic, or fixed vector of statistics, in arm $m$,
define the adjusted run value
\begin{equation}
    Y^{\mathrm{adj}}_{m,s}
    = Y_{m,s} - \theta_m^\top (X_{m,s} - \mu_{X,m}),
    \label{eq:adj}
\end{equation}
where $\mu_{X,m} = \mathbb{E}[X_{m,s}]$. If $\mu_{X,m}$ is known, the centered
covariate has mean zero and
$\mathbb{E}[Y^{\mathrm{adj}}_{m,s}] = \mu_m$ for any fixed $\theta_m$. The
coefficient affects variance, not the target. For a single covariate with the
population-optimal coefficient, the variance reduction (VR) is
\begin{equation}
    \mathrm{VR}_m = 1 -
    \frac{\mathrm{Var}(Y^{\mathrm{adj}}_{m})}{\mathrm{Var}(Y_m)}
    = \rho_{X_mY_m}^2.
    \label{eq:vr}
\end{equation}
In finite samples, the realized VR is smaller than $\rho^2$ because $\theta_m$
must be estimated; the cost scales roughly as $1/n_{\mathrm{train}}$
(Section~\ref{sec:selection_risk}).
Thus, a training statistic that explains run-level variation can reduce the
standard error of $\mu_m$ and, when applied independently in both arms, reduce
the standard error of $\Delta$.

The adjustment must be arm-specific. A pooled regression of outcomes on model
identity and co-produced training statistics can remove part of the model
difference, because early validation loss or gradient norms may differ between
models precisely because the models train differently. We therefore fit
the adjustment separately within each model and combine the two arm-specific
standard errors only after adjustment.

\subsection{Covariate Selection Risk}
\label{sec:selection_risk}

If the covariate is chosen after screening many candidates, the apparent
correlation can be optimistic and the realized VR can be negative. Intuitively,
the covariate that looks most correlated on a small training sample may not
remain correlated on held-out runs, causing the adjustment to add noise rather
than remove it. The estimation penalty scales as
$\sim p / n_{\mathrm{train}}$, where $p$ is the effective number of
covariates. For a single selected covariate, the nominal $p = 1$, but the
implicit search over hundreds of candidates inflates the effective degrees of
freedom. This risk motivates out-of-fold estimation
(Section~\ref{sec:crossfit}) and is analyzed empirically in
Section~\ref{sec:selection_failure}.

\subsection{Why Co-Produced Covariates Need Care}
\label{sec:validity}

In standard regression adjustment for A/B tests, covariates are measured
before the treatment, which separates covariate measurement from treatment
assignment. This separation helps prevent a covariate adjustment from
absorbing part of the treatment effect~\citep{rosenbaum1984consequences}.
Training-log covariates
are different: they are produced by the same stochastic run whose final
accuracy is being evaluated.

Two design choices make the adjustment scientifically interpretable in our
setting. First, we use covariates only within the arm that produced them. This
keeps the target for each arm as the expected final accuracy of that model,
rather than a performance contrast after controlling for a shared
post-treatment variable. Second, we evaluate each proposed training-log
adjustment by its out-of-fold variance reduction and by coverage diagnostics
based on resampling from the 50 runs we actually trained, rather than assuming
that a reduced residual variance automatically yields a valid hypothesis test.

If the training-log statistic and population covariate mean were fixed in
advance, the centering in Equation~\ref{eq:adj} would preserve the arm mean.
In practice, both the coefficient and the covariate mean are estimated from the
same limited run budget. Cross-fitting reduces direct overfitting, but the
results still report both variance reduction and interval coverage to verify
the adjustment empirically.

\subsection{Cross-Fitted Evaluation}
\label{sec:crossfit}

The population expression above assumes that $\theta_m$ and $\mu_{X,m}$ are
known. In practice, both must be estimated from the same runs used for
evaluation. This creates a risk: the adjustment might reduce apparent variance
by memorizing run-level noise rather than capturing stable structure. We use
$K$-fold cross-fitting~\citep{chernozhukov2018dml} with $K=5$ throughout. Runs are partitioned into
$K$ folds, and for run $s$ in fold $f(s)$, nuisance parameters are estimated
from the complement:
\begin{equation}
    \tilde{Y}_{m,s} = Y_{m,s} - \hat{\theta}_{m,-f(s)}^\top
    \bigl(X_{m,s} - \hat{\mu}_{X,m,-f(s)}\bigr).
\end{equation}
For single-arm diagnostics, we compare the sample variance of the cross-fitted
adjusted outcomes to the raw outcome variance. For pairwise comparisons, we
recenter adjusted outcomes within each arm to keep the reported performance
difference equal to the raw difference, then compute a Welch-style interval
using the adjusted within-arm variances. This conservative reporting choice
separates the model ranking from the variance estimate: covariates can change
the reported uncertainty, but not the reported mean performance difference.

Because adjusted outcomes within a fold share fitted coefficients, and because
we recenter to preserve the raw point estimate, Section~\ref{sec:coverage}
checks adjusted-interval coverage by repeatedly resampling from the 50 runs we
actually trained.

Cross-fitting prevents the coefficient from directly memorizing held-out
runs. However, if the covariate is \emph{selected} from a large pool using the
same runs, the selection step must also be kept out of fold to avoid choosing a
statistic that correlates with noise in the evaluation runs. As we show in
Section~\ref{sec:selection_failure}, this selection problem is the primary practical
bottleneck.

\subsection{Pairwise Intervals}
\label{sec:power_method}

For two models A and B, arm-specific adjustment gives estimated within-arm
standard deviations $s_{\mathrm{adj},A}$ and $s_{\mathrm{adj},B}$. We use the
raw mean difference as the point estimate and compute the adjusted standard
error
\begin{equation}
    \widehat{\mathrm{SE}}_{\mathrm{adj}}(\hat{\Delta}) =
    \sqrt{\frac{s_{\mathrm{adj},A}^2}{n_A} +
          \frac{s_{\mathrm{adj},B}^2}{n_B}}.
    \label{eq:se_diff}
\end{equation}
The corresponding confidence interval uses the Welch degrees of freedom. When
both adjusted variances are smaller than their raw counterparts, the interval
for the performance difference tightens. If one arm has a poor covariate, the
interval can widen. The minimum detectable effect size (MDE) at power $1-\beta$ and
significance $\alpha$ is
\begin{equation}
    \mathrm{MDE} \approx (z_{1-\alpha/2} + z_{1-\beta}) \cdot
    \sqrt{\frac{s_{\mathrm{adj},A}^2 + s_{\mathrm{adj},B}^2}{n}},
    \label{eq:mde}
\end{equation}
for equal run counts (the normal quantiles are asymptotic; our intervals
use the Welch $t$ approximation). Thus,
interval tightening also corresponds to improved test sensitivity when
uncertainty is the limiting factor.

We now describe the experimental design used to test these claims empirically.

\section{Experimental Setup}
\label{sec:setup}

\subsection{Design}

We conduct a $3 \times 3$ factorial experiment: three architectures $\times$
three datasets, with 50 runs per cell (450 runs total). All experiments run on
NVIDIA L4 GPUs with bfloat16 mixed precision. Sources of randomness (Python,
NumPy, PyTorch, CUDA, DataLoader workers) are seeded deterministically.
The primary outcome is \textbf{final-epoch test accuracy}, fixed before the
covariate analysis. Using the best-validation-accuracy checkpoint would make
validation metrics part of the outcome selection, preventing their use as
covariates. With a fixed checkpoint, validation logs can serve as covariates
without leakage. Across the nine cells,
baseline run standard deviations range from 0.13\,pp (ConvNeXt-Tiny / CIFAR-10)
to 0.69\,pp (ConvNeXt-Tiny / Tiny-ImageNet); full results are in
Appendix~\ref{app:variability}.

\subsection{Architectures and Recipes}

\textbf{ResNet-18}~\citep{he2016deep} with a stem adapted for
$32\!\times\!32$ inputs (stride-1 $3\!\times\!3$ conv, no max-pool), trained
with SGD (lr 0.1, cosine schedule, 100 to 200 epochs depending on dataset).
\textbf{ViT-Tiny}~\citep{touvron2021deit} (embed dim 192, 12 layers, 3 heads,
patch size 4 on CIFAR, 8 on Tiny-ImageNet) and
\textbf{ConvNeXt-Tiny}~\citep{liu2022convnet} (resolution-adapted stems), both
trained with AdamW (cosine schedule, RandAugment, Mixup, CutMix, 300 epochs).
ViT-Tiny applies gradient clipping at 1.0; ConvNeXt-Tiny omits clipping and
EMA to preserve run-specific variance signals.

\subsection{Datasets}

CIFAR-10 and CIFAR-100~\citep{krizhevsky2009learning}
($32\!\times\!32$, 50k train / 10k test) and
Tiny-ImageNet~\citep{le2015tiny}
($64\!\times\!64$, 100k train / 10k test, 200 classes). From each training
set, 5,000 images are held out as a fixed validation split (split seed 2026).

\subsection{Covariates}

Every epoch logs train/val/test loss and accuracy and gradient norm statistics.
The first 1,000 training steps log per-batch loss, accuracy, gradient norm,
and parameter norm. At completion, $\sim$200 summary statistics are computed
per run (prefix means, standard deviations, slopes, etc.). We restrict
covariates to the first third of training and exclude test-derived statistics.
Candidates are organized into five families:
\begin{itemize}[nosep]
    \item \textbf{Validation snapshots}: val\_acc and val\_loss at early epochs
    (roughly the first third of training).
    \item \textbf{Training-loss summaries}: epoch losses, per-batch loss
    summaries, and first-step loss statistics.
    \item \textbf{Training-accuracy summaries}: epoch training accuracy,
    per-batch accuracy summaries, and first-step accuracy statistics.
    \item \textbf{Gradient-norm summaries}: per-epoch and per-step gradient
    norms and their derived statistics.
    \item \textbf{Parameter-norm summaries}: per-epoch and per-step parameter
    norms, learning rate, and their derived statistics.
\end{itemize}
\noindent The total candidate count ranges from 672 (ResNet-18 / CIFAR-10) to
1,719 (ViT-Tiny / CIFAR-10). Two training details affect covariate
interpretation: for ViT-Tiny, gradient norms are logged post-clipping (capped at
$\leq 1.0$), limiting their informativeness; and for ViT-Tiny and ConvNeXt-Tiny,
training accuracy is computed against hard labels while training uses mixed soft
targets from Mixup/CutMix.

\subsection{Choosing Training-Log Covariates}

The primary pairwise experiments adjust each arm using validation loss at the
first-third epoch. This statistic is chosen before comparing the empirical
covariance structure across the candidate pool and is motivated by prior work
on learning-curve prediction~\citep{domhan2015speeding}. It provides a
prospective baseline for testing whether training logs can reduce uncertainty
without searching over hundreds of candidates.

We also evaluate two data-driven alternatives to diagnose whether stronger
adjustments can be obtained from the log pool. \textbf{Single-best OLS} asks
whether directly searching the logs can find a useful statistic: within each
training fold, it chooses the candidate with the largest $\rho^2$ with final
accuracy, then applies the fitted coefficient to held-out runs. This procedure
is intuitive, but it uses the outcome during selection and can overfit when the
candidate pool is large. \textbf{PCA ($k=1$)} takes a more restricted route: it
replaces each family of training-log statistics with its first principal
component and fits OLS on those components. PCA does not use final accuracy
when constructing the components within a fold, while single-best OLS does use
final accuracy to choose among candidates. Comparing the two helps separate
useful signal in the logs from the noise introduced by searching over many
candidate covariates.

\section{Results}
\label{sec:results}

We report results for all nine configurations in the $3\times3$ design.
Early-training covariates often reduce variance, but choosing among many
training-log statistics can introduce more estimation error than it removes.
We first check interval calibration and null behavior, then evaluate pairwise
model comparisons, and finally use single-arm variance reduction,
covariate-selection diagnostics, and run-budget analysis to explain when
adjustment helps or hurts.

\subsection{Interval Calibration}
\label{sec:coverage}

Before interpreting variance reduction or interval width, we assess whether
adjusted intervals become misleadingly narrow. In standard A/B testing,
covariates are measured before treatment, which separates covariate measurement
from treatment assignment. Training-log covariates are co-produced with the
outcome, so this separation does not hold. If adjustment distorts coverage,
tighter intervals would reflect false confidence rather than improved
precision.

We test this with a subsampling diagnostic. For each model-dataset cell and
run budget $n \in \{10, 15, 20, 30, 50\}$, we repeatedly draw a random subset
of $n$ runs from the full pool of 50 (without replacement within each draw),
build a cross-fitted 95\% confidence interval from that subset, and check
whether it contains the 50-run mean. We repeat this 5,000 times per setting.
Under-coverage in this finite-run diagnostic would indicate that adjustment is
making intervals too narrow; coverage at or above 95\% is a minimal calibration
check.

\begin{table}[t]
\caption{Interval coverage (\%) across run budgets for CIFAR-100.}
\label{tab:coverage}
\vskip 0.05in
{\footnotesize Raw = unadjusted; Pre = validation loss at first-third epoch.}
\vskip 0.1in
\begin{center}
\begin{small}
\begin{tabular}{@{}lcccccc@{}}
\toprule
 & \multicolumn{2}{c}{\textbf{ResNet}} & \multicolumn{2}{c}{\textbf{ViT-T}}
 & \multicolumn{2}{c}{\textbf{ConvNeXt}} \\
\cmidrule(lr){2-3} \cmidrule(lr){4-5} \cmidrule(lr){6-7}
$n$ & Raw & Pre & Raw & Pre & Raw & Pre \\
\midrule
10 & 96.6 & 97.5 & 96.7 & 96.8 & 97.2 & 96.6 \\
15 & 97.9 & 98.5 & 98.2 & 97.5 & 97.9 & 97.2 \\
20 & 98.6 & 99.1 & 98.9 & 98.3 & 98.7 & 98.3 \\
30 & 99.7 & 99.8 & 99.8 & 99.7 & 99.9 & 99.8 \\
50 & 100 & 100 & 100 & 100 & 100 & 100 \\
\bottomrule
\end{tabular}
\end{small}
\end{center}
\vskip -0.15in
\end{table}

Table~\ref{tab:coverage} shows results for CIFAR-100; the same diagnostic on
the other datasets also shows no under-coverage. All entries are at or above
95\%, suggesting that adjustment does not make intervals anti-conservative
when we resample from the 50 runs we actually trained. Coverage exceeds 95\%
at small $n$ because the $t_{n-1}$ quantile is conservative with few
observations. The $n=50$ row is a deterministic limit case: drawing all 50
runs yields one subset, which covers its own mean by construction. We include
it as an implementation check. Because the 5,000 subsamples are drawn from the
same 50 runs, the coverage indicators are positively correlated; the decimal
precision should not be over-interpreted.

\subsection{Null Calibration}
\label{sec:null}

The coverage diagnostic above checks whether adjusted intervals cover the
mean of the 50 runs we actually trained within each model-dataset cell, but it
does not test behavior under a zero model difference. To probe Type~I behavior
under a known-zero contrast within the observed run pool, we construct a
synthetic null:
for each cell, we randomly split the 50 runs from the \emph{same} model into
two fake arms of 25. The two arms are exchangeable under this construction, so
the null hypothesis of equal expected performance holds even though any
particular split can have a nonzero sample mean difference. We then compute
the adjusted Welch interval and check whether it covers zero. Across 5,000
random splits per cell, adjusted coverage averages 94.8\% (range 92.9\% to
95.9\%), close to the nominal 95\%. The raw interval averages 95.1\%.
Adjusted rejection rates average 5.2\%, compared to 4.9\% raw. One cell
(CIFAR-10 / ViT-Tiny) shows elevated rejection at 7.1\%. Because the random
splits share runs and we examine nine cells, we treat this elevation as a
diagnostic warning rather than a formal significance claim. The pattern
suggests that arm-specific validation-loss adjustment does not systematically
inflate rejection rates, but that null calibration should be reported rather
than assumed, especially for cells with strong covariate signal.

\subsection{Pairwise Model Comparisons}
\label{sec:pairwise}

After the calibration checks, we evaluate the main inferential target:
confidence intervals for model-performance differences. We use the fixed
validation-loss adjustment from Section~\ref{sec:setup} and fit it
independently per arm with cross-fitting (2,000 subsamples per run budget).
Table~\ref{tab:pairwise} reports the percent change in the 95\% Welch
interval half-width; positive values mean the adjusted interval is narrower.

\begin{table}[t]
\caption{Pairwise CI diagnostics from arm-specific validation-loss adjustment.}
\label{tab:pairwise}
\vskip 0.05in
{\footnotesize $\Delta$CI = percent change in half-width (positive = narrower).
HW = absolute half-width in pp at $n=50$. Cov = minimum adjusted coverage over
$n \in \{10,15,20,30,50\}$ when resampling from the trained runs (\%).}
\vskip 0.1in
\begin{center}
\resizebox{\columnwidth}{!}{
\begin{small}
\begin{tabular}{@{}lrrrrrr@{}}
\toprule
 & & \multicolumn{2}{c}{$\Delta$CI (\%)} &
   \multicolumn{2}{c}{HW (pp), $n$=50} & Cov \\
\cmidrule(lr){3-4} \cmidrule(lr){5-6} \cmidrule(lr){7-7}
\textbf{Pair} & \textbf{Diff} & $n$=10 & $n$=50 & Raw & Adj & (\%) \\
\midrule
C10 / R-V   & 1.00 & $-$5.8  & 4.5  & 0.0758 & 0.0724 & 97.2 \\
C10 / R-CN  & $-$1.93 & $-$12.4 & $-$1.4 & 0.0548 & 0.0555 & 98.2 \\
C10 / V-CN  & $-$2.93 & $-$4.0  & 5.0  & 0.0744 & 0.0706 & 97.2 \\
\midrule
C100 / R-V  & 6.98 & $-$2.0  & 6.6  & 0.1457 & 0.1360 & 97.6 \\
C100 / R-CN & $-$2.30 & $-$5.0  & 3.9  & 0.1153 & 0.1108 & 97.2 \\
C100 / V-CN & $-$9.28 & $-$0.1  & 8.8  & 0.1584 & 0.1445 & 97.5 \\
\midrule
TIN / R-V   & 9.88 & $-$6.6  & 2.3  & 0.1453 & 0.1419 & 97.4 \\
TIN / R-CN  & $-$2.39 & $-$7.5  & 1.9  & 0.2083 & 0.2044 & 97.2 \\
TIN / V-CN  & $-$12.27 & $-$6.4  & 2.7  & 0.2293 & 0.2232 & 97.5 \\
\bottomrule
\end{tabular}
\end{small}
}
\end{center}
\vskip -0.15in
\end{table}

The pairwise results show conditional gains. At $n=50$, adjusted intervals
narrow in eight of nine model pairs, with reductions from 1.9\% to
8.8\% among the successful pairs. The only exception is CIFAR-10 /
ResNet-18 vs.\ ConvNeXt-Tiny ($-$1.4\%), where the later single-arm analysis
shows that validation-loss adjustment is not useful in either arm. The absolute
magnitudes are modest: the largest improvement is CIFAR-100 / ViT-Tiny vs.\
ConvNeXt-Tiny, where the half-width shrinks from 0.158\,pp to 0.144\,pp.
These gains require a sufficient run budget; at very small budgets (e.g.,
$n=10$), the cost of estimating the covariate coefficient systematically
exceeds the variance removed, and the adjusted interval widens across all
pairs. Adjustment pays off only when the run budget is large enough for stable
coefficient estimation. When resampling from the trained runs, adjusted
pairwise coverage remains $\geq$97.2\% across all pairs and run budgets,
consistent with no clear anti-conservatism in this diagnostic.

Since MDE scales with the same standard error (Equation~\ref{eq:mde}), these
interval reductions correspond to higher sensitivity in comparisons where
uncertainty is practically relevant.

\subsection{Single-Arm Variance Reduction}
\label{sec:vr}

To explain why the pairwise gains are conditional, we measure single-arm
variance reduction (VR) for the three adjustments defined in
Section~\ref{sec:setup}: validation loss at the first-third epoch,
\textbf{single-best OLS}, and \textbf{PCA ($k$=1)}.
We define $\mathrm{VR} = 1 - \mathrm{Var}(\tilde{Y}) / \mathrm{Var}(Y)$,
where variances are computed over cross-fitted adjusted and raw outcomes
using repeated 5-fold cross-fitting (1,000 repeats). Negative VR means the
adjustment \emph{increases} variance: the estimation cost of fitting
$\theta$ exceeds the variance removed.

\begin{table}[t]
\caption{Variance reduction (\%) by adjustment.}
\label{tab:results}
\vskip 0.05in
{\footnotesize Parentheses in OLS and PCA columns: percentage of repeated
cross-fitting splits with VR $<$ 0. Bold = best VR per cell. Pre = validation
loss at first-third epoch; OLS = single-best out-of-fold selector; PCA = first
PC per family.}
\vskip 0.1in
\begin{center}
\resizebox{\columnwidth}{!}{
\begin{small}
\begin{tabular}{@{}lrrr@{}}
\toprule
\textbf{Configuration} & \textbf{Pre} & \textbf{OLS} & \textbf{PCA} \\
\midrule
C10 / ResNet-18 & $-$2.9 & $-$35.4 (100) & $-$1.8 (50) \\
C10 / ViT-Tiny & 13.6 & $-$38.1 (100) & \textbf{23.1 (0)} \\
C10 / ConvNeXt-T & $-$2.1 & \textbf{13.3 (19)} & 0.5 (42) \\
\midrule
C100 / ResNet-18 & $-$5.8 & $-$41.1 (100) & $-$17.6 (100) \\
C100 / ViT-Tiny & 17.8 & $-$18.6 (97) & \textbf{21.7 (0)} \\
C100 / ConvNeXt-T & 14.5 & \textbf{34.1 (2)} & 9.7 (7) \\
\midrule
TIN / ResNet-18 & $-$2.1 & $-$28.5 (100) & $-$12.4 (100) \\
TIN / ViT-Tiny & \textbf{7.0} & $-$4.5 (58) & 4.8 (18) \\
TIN / ConvNeXt-T & 4.7 & 30.5 (3) & \textbf{34.0 (0)} \\
\bottomrule
\end{tabular}
\end{small}
}
\end{center}
\vskip -0.15in
\end{table}

\paragraph{Fixed validation-loss adjustment.} Validation loss at the first-third
epoch gives positive VR for ViT-Tiny on all three datasets (7.0 to 17.8\%)
and for ConvNeXt-Tiny on two of three, but is uniformly negative for
ResNet-18. Because no selection is involved, these results isolate covariate
quality from selection noise: the fixed validation-loss adjustment
helps some architecture and training-recipe combinations but not others.

\paragraph{Data-driven selection.} The single-best OLS selector produces
negative VR in six of nine cells. The failures are not small: ViT-Tiny on
CIFAR-10 has $\rho^2 = 23\%$ for its best early validation covariate, yet
the selector yields $-38.1\%$ VR. This pattern identifies covariate selection
as the main empirical failure mode in our study.

\paragraph{PCA.} PCA ($k$=1) avoids outcome-based selection and recovers
positive VR for ViT-Tiny on all datasets and for Tiny-ImageNet /
ConvNeXt-Tiny, outperforming both alternatives in three of nine cells.
Increasing to $k=2$ or $k=3$ PCs per family degrades performance, consistent
with overfitting when the number of fitted coefficients grows relative to the
run budget. PCA's stability comes from the fact that the first principal
component is a deterministic function of the covariate matrix within each
fold, avoiding outcome-based discrete selection entirely.

\begin{figure}[t]
    \centering
    \includegraphics[width=\columnwidth]{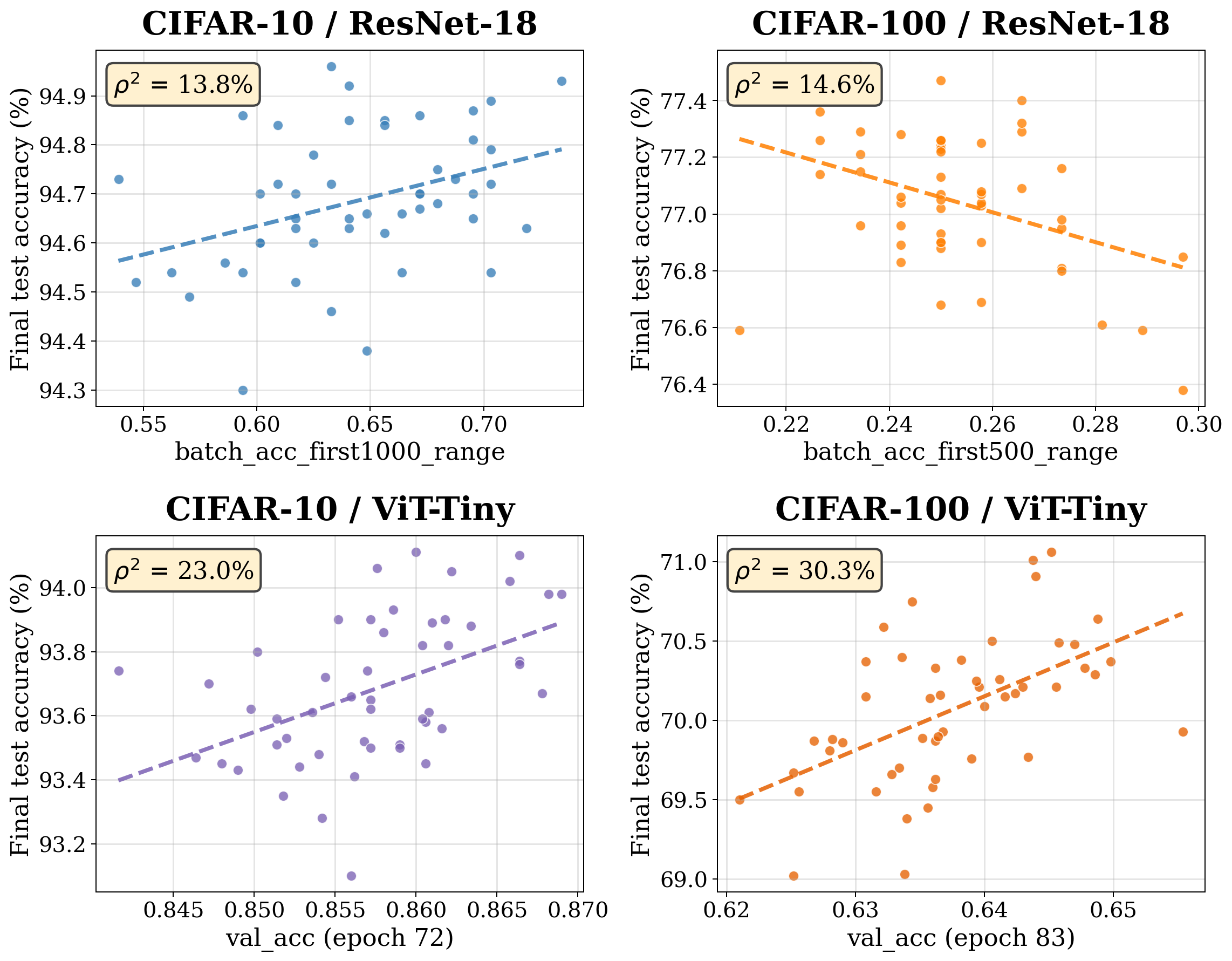}
    \caption{Covariate-outcome scatter for four cells. Each point is one
    run.}
    \label{fig:scatter}
    {\footnotesize ViT-Tiny (bottom) shows stronger first-third relationships
    ($\rho^2 = 23.0\%$ and $30.3\%$); ResNet-18 (top) shows weaker
    relationships ($\rho^2 = 13.8\%$ and $14.6\%$).}
\end{figure}

\paragraph{Post hoc evidence of signal.} To assess whether useful covariates
exist even when selection fails, we run exploratory scans using all 50 runs
for selection (but cross-fitting the coefficient). Every cell has at least one
first-third covariate with positive VR (11.7\% to 44.3\%). The examples are
heterogeneous: ViT-Tiny's best covariates are early validation snapshots
($\rho^2$ 23 to 30\%), whereas ResNet-18's are gradient and batch-loss
summaries ($\rho^2 \leq 18.8\%$). These discovery results suggest that signal
exists but is difficult to extract reliably at current run budgets.
Figure~\ref{fig:scatter} illustrates the contrast using first-third
covariates: ViT-Tiny shows stronger linear covariate-outcome relationships,
while ResNet-18 shows weaker, noisier correlations.

\subsection{Why Data-Driven Selection Fails}
\label{sec:selection_failure}

The discrepancy between post hoc signal and nested-selection performance
suggests that the main difficulty is not signal absence, but finite-sample
selection noise. We examine this mechanism with a family-level diagnostic.

\paragraph{Diagnostic setup.} We split candidates into five families
(validation, training-loss, training-accuracy, gradient-norm, parameter-norm)
and compare two adjustment procedures within each:
\begin{itemize}[nosep]
    \item \emph{Post hoc best}: use all 50 runs to identify the
    highest-$R^2$ covariate, then cross-fit only the coefficient.
    This is biased upward because the selection step uses evaluation runs.
    \item \emph{Nested selector}: use only the $\sim$40 training-fold runs
    for both selection and coefficient estimation.
\end{itemize}

\begin{figure}[t]
    \centering
    \includegraphics[width=0.85\columnwidth]{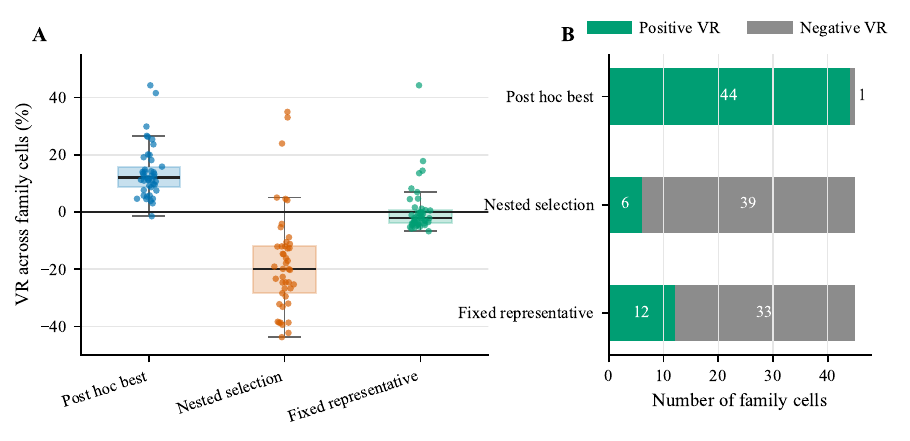}
    \caption{Family-level selection diagnostic.}
    \label{fig:selection_gap}
    {\footnotesize Post hoc best covariates (using all 50 runs for selection)
    are usually positive; nested within-fold selection shifts VR below zero.
    The gap is selection noise.}
\end{figure}

\paragraph{Results.} The post hoc best is positive in 44 of 45 family-cell
combinations; the nested selector is negative in 39 of 45
(Figure~\ref{fig:selection_gap}). Even within a single family, the candidate
count is large relative to the training-fold size, so the nested selector
overfits to fold-specific noise.

\paragraph{Interpretation.} The estimation penalty scales as
$\sim p/n_{\mathrm{train}}$, where $p$ is the number of covariates. For a
single selected covariate, the nominal $p=1$, but searching over hundreds of
candidates inflates the effective degrees of freedom. The ConvNeXt-Tiny cells
provide independent corroboration: the nested selector gives positive VR in
all three ConvNeXt-Tiny configurations, and the selected covariates are
interpretable early-training summaries. Two regimes emerge: when signal is
distributed broadly across a family, PCA captures the dominant relationship
without selection; when signal is concentrated in a few covariates (as in
ConvNeXt-Tiny), single-best OLS identifies them directly.

\subsection{Variance Reduction vs.\ Run Budget}
\label{sec:budget}

\begin{figure}[t]
    \centering
    \includegraphics[width=0.85\columnwidth]{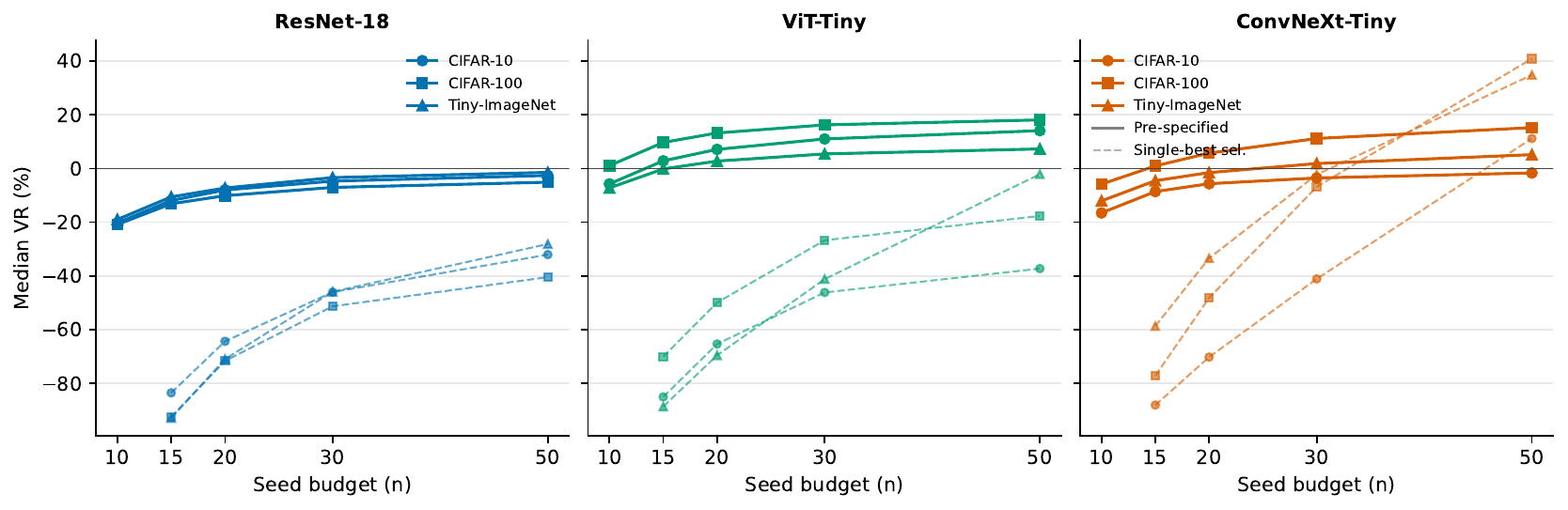}
    \caption{Median VR vs.\ run budget.}
    \label{fig:vr_vs_n}
    {\footnotesize Solid: fixed validation-loss adjustment; dashed:
    single-best selector. The fixed adjustment degrades more smoothly; automatic
    selection remains brittle at all tested budgets.}
\end{figure}

At smaller run budgets, fixed validation-loss adjustment degrades more gracefully than
data-driven selection, which is not reliably positive below $n=50$ in these
experiments (Figure~\ref{fig:vr_vs_n}). For ViT-Tiny, validation loss at the
first-third epoch gives positive median VR from $n = 15$ onward (+2.8\% at $n=15$,
+14.1\% at $n=50$ on CIFAR-10). At $n < 15$, the cost of estimating
$\theta$ from too few runs exceeds the variance removed. This threshold is
consistent with the heuristic that realized VR is
approximately $\rho^2 - 1/n_{\mathrm{train}}$: a covariate needs
$\rho^2 > 1/n_{\mathrm{train}}$ to pay for itself.

\section{Discussion}
\label{sec:discussion}

The main lesson is modest but useful: training logs can make model comparisons
more precise when the adjustment is chosen before the comparison or avoids
outcome-based selection. At the largest run budget we study, the fixed early
validation-loss adjustment narrows eight of nine pairwise confidence
intervals. The largest single-arm gains come from ConvNeXt-Tiny, where
single-best OLS reaches 34.1\% variance reduction and PCA reaches 34.0\%.
The 34.1\% reduction is equivalent to 75.9 effective runs from a 50-run
budget. This effective-run calculation is a diagnostic scale, not a
recommendation to train that many repeats in routine comparisons.

The failure cases matter just as much. The log pool contains useful
statistics, as shown by the post hoc family scans, but choosing the most
correlated statistic within each training fold usually adds more noise than it
removes. In the family diagnostic, the post hoc best statistic is positive in
44 of 45 family-cell combinations, while nested within-fold selection is
negative in 39 of 45. In these experiments, the bottleneck is estimating and
selecting the adjustment from limited runs, not simply a lack of signal in the
logs.

The useful signal also varies across models. ConvNeXt-Tiny gives the largest
gains, ViT-Tiny benefits from the fixed validation-loss adjustment, and
ResNet-18 shows little benefit under the adjustments tested here. A
training-log statistic that helps one model should not be assumed to help
another. Pairwise comparisons should therefore adjust each arm separately and
show the raw interval alongside the adjusted interval.

Reports of adjusted comparisons should include the raw mean difference and
confidence interval, the adjusted interval, the exact training-log statistic
or summary used in each arm, and the out-of-fold variance reduction, including
how often it is negative across cross-fitting splits. If the statistic was
chosen from a large candidate pool after looking at outcome correlations, the
adjusted interval is better treated as exploratory.

\section{Limitations}
\label{sec:limitations}

The adjusted results have an important inferential limit. Training-log
covariates are recorded during the same stochastic runs that produce the final
accuracies. As a result, the adjusted intervals should be read as empirically
checked precision estimates, not as distribution-free finite-sample guarantees.
A stronger guarantee would require the adjustment to be fixed in advance and
the covariate mean to be known or estimated independently. For this reason, we
keep the raw interval as the baseline, report adjusted intervals separately,
and check the adjusted intervals by resampling from the 50 runs we actually
trained and by using the synthetic-null diagnostic. The synthetic null is
broadly consistent with nominal rejection but has one elevated cell.

The experiments use CIFAR-10, CIFAR-100, and Tiny-ImageNet. Larger datasets,
larger models, and different run budgets may change both the available
covariate signal and the cost of estimating the adjustment. Because
architecture and training recipe vary together, we treat the differences across
ResNet-18, ViT-Tiny, and ConvNeXt-Tiny as configuration-level evidence rather
than as an isolated architecture effect. The adjustment model is also linear.
Nonlinear adjustment might recover additional signal, but it would increase
the risk of overfitting at the run budgets studied here.

Finally, the outcome is final-epoch test accuracy. Validation metrics can be
used as training-log covariates in this setting because validation accuracy
does not choose the reported checkpoint. If validation performance determines
checkpoint selection or early stopping, validation-derived covariates would
introduce leakage and should not be used for the adjustment. Sequential
evaluation is another open direction: the fixed-sample interval studied here
could be embedded within an always-valid confidence
sequence~\citep{johari2022always}, where variance reduction from covariate
adjustment would translate to faster stopping once the adjusted interval
reaches a target width.

\section{Conclusion}
\label{sec:conclusion}

Training logs are usually treated as optimization diagnostics. This paper asks
whether they can also make statistical comparisons between stochastically
trained models more precise. Arm-specific covariate adjustment leaves the raw
mean difference unchanged, but can reduce the standard error of that difference
when each arm has a stable training-log statistic. In our 450-run study, a
fixed early validation-loss adjustment narrows most pairwise confidence
intervals at the largest run budget we study. PCA summaries of training-log
families also show that some log families contain additional single-arm
variance signal.

The results should not be read as recommending dozens of repeated runs for
routine model comparisons. They show a different point: training logs can
reduce uncertainty once enough repeated runs are available to estimate the
adjustment stably. Selecting the most correlated statistic from a large log
pool usually worsens variance after cross-fitting, even when useful statistics
exist in hindsight. Training logs can help make model comparisons
more precise, but the practical bottleneck is reliable covariate selection
under limited run budgets.

\section*{Impact Statement}
This paper presents work whose goal is to advance the methodology of
evaluating machine learning systems. There are many potential societal
consequences of our work, none of which we feel must be specifically
highlighted here.

\bibliography{references}

\begin{thebibliography}{22}
\providecommand{\natexlab}[1]{#1}
\providecommand{\url}[1]{\texttt{#1}}
\expandafter\ifx\csname urlstyle\endcsname\relax
  \providecommand{\doi}[1]{doi: #1}\else
  \providecommand{\doi}{doi: \begingroup \urlstyle{rm}\Url}\fi

\bibitem[Bouthillier et~al.(2021)Bouthillier, Delaunay, Bronzi, Trofimov,
  et~al.]{bouthillier2021accounting}
Bouthillier, X., Delaunay, P., Bronzi, M., Trofimov, A., et~al.
\newblock Accounting for variance in machine learning benchmarks.
\newblock In \emph{Proceedings of Machine Learning and Systems}, volume~3, pp.\
   747--769, 2021.

\bibitem[Chernozhukov et~al.(2018)Chernozhukov, Chetverikov, Demirer, Duflo,
  Hansen, Newey, and Robins]{chernozhukov2018dml}
Chernozhukov, V., Chetverikov, D., Demirer, M., Duflo, E., Hansen, C., Newey,
  W., and Robins, J.
\newblock Double/debiased machine learning for treatment and structural
  parameters.
\newblock \emph{The Econometrics Journal}, 21\penalty0 (1):\penalty0 C1--C68,
  2018.

\bibitem[Dehghani et~al.(2021)Dehghani, Tay, Gritsenko, Zhao, Houlsby, Diaz,
  Metzler, and Vinyals]{dehghani2021benchmark}
Dehghani, M., Tay, Y., Gritsenko, A.~A., Zhao, Z., Houlsby, N., Diaz, F.,
  Metzler, D., and Vinyals, O.
\newblock The {Benchmark} {Lottery}.
\newblock \emph{arXiv preprint arXiv:2107.07002}, 2021.

\bibitem[Deng et~al.(2013)Deng, Xu, Kohavi, and Walker]{deng2013cuped}
Deng, A., Xu, Y., Kohavi, R., and Walker, T.
\newblock Improving the sensitivity of online controlled experiments by
  utilizing pre-experiment data.
\newblock In \emph{Proceedings of the Sixth ACM International Conference on Web
  Search and Data Mining}, pp.\  123--132, 2013.

\bibitem[Dodge et~al.(2020)Dodge, Ilharco, Schwartz, Farhadi, Hajishirzi, and
  Smith]{dodge2020fine}
Dodge, J., Ilharco, G., Schwartz, R., Farhadi, A., Hajishirzi, H., and Smith,
  N.
\newblock Fine-tuning pretrained language models: Weight initializations, data
  orders, and early stopping.
\newblock \emph{arXiv preprint arXiv:2002.06305}, 2020.

\bibitem[Domhan et~al.(2015)Domhan, Springenberg, and
  Hutter]{domhan2015speeding}
Domhan, T., Springenberg, J.~T., and Hutter, F.
\newblock Speeding up automatic hyperparameter optimization of deep neural
  networks by extrapolation of learning curves.
\newblock In \emph{Proceedings of the Twenty-Fourth International Joint
  Conference on Artificial Intelligence}, pp.\  3460--3468, 2015.

\bibitem[Freedman(2008)]{freedman2008regression}
Freedman, D.~A.
\newblock On regression adjustments to experimental data.
\newblock \emph{Advances in Applied Mathematics}, 40\penalty0 (2):\penalty0
  180--193, 2008.

\bibitem[Gelman et~al.(2013)Gelman, Carlin, Stern, Dunson, Vehtari, and
  Rubin]{gelman2013bayesian}
Gelman, A., Carlin, J.~B., Stern, H.~S., Dunson, D.~B., Vehtari, A., and Rubin,
  D.~B.
\newblock \emph{Bayesian Data Analysis}.
\newblock Chapman and Hall/CRC, 3rd edition, 2013.

\bibitem[He et~al.(2016)He, Zhang, Ren, and Sun]{he2016deep}
He, K., Zhang, X., Ren, S., and Sun, J.
\newblock Deep residual learning for image recognition.
\newblock In \emph{Proceedings of the IEEE Conference on Computer Vision and
  Pattern Recognition}, pp.\  770--778, 2016.

\bibitem[Henderson et~al.(2018)Henderson, Islam, Bachman, Pineau, Precup, and
  Meger]{henderson2018deep}
Henderson, P., Islam, R., Bachman, P., Pineau, J., Precup, D., and Meger, D.
\newblock Deep reinforcement learning that matters.
\newblock In \emph{Proceedings of the AAAI Conference on Artificial
  Intelligence}, volume~32, 2018.

\bibitem[Johari et~al.(2022)Johari, Koomen, Pekelis, and
  Walsh]{johari2022always}
Johari, R., Koomen, P., Pekelis, L., and Walsh, D.
\newblock Always valid inference: Continuous monitoring of a/b tests.
\newblock \emph{Operations Research}, 70\penalty0 (3):\penalty0 1806--1821,
  2022.

\bibitem[Krizhevsky(2009)]{krizhevsky2009learning}
Krizhevsky, A.
\newblock Learning multiple layers of features from tiny images.
\newblock \emph{Technical report, University of Toronto}, 2009.

\bibitem[Le \& Yang(2015)Le and Yang]{le2015tiny}
Le, Y. and Yang, X.
\newblock Tiny imagenet visual recognition challenge.
\newblock \emph{CS 231N}, 7\penalty0 (7):\penalty0 3, 2015.

\bibitem[Lin(2013)]{lin2013agnostic}
Lin, W.
\newblock Agnostic notes on regression adjustments to experimental data:
  {R}eexamining {F}reedman's critique.
\newblock \emph{The Annals of Applied Statistics}, 7\penalty0 (1):\penalty0
  295--318, 2013.

\bibitem[Liu et~al.(2022)Liu, Mao, Wu, Feichtenhofer, Darrell, and
  Xie]{liu2022convnet}
Liu, Z., Mao, H., Wu, C.-Y., Feichtenhofer, C., Darrell, T., and Xie, S.
\newblock A {ConvNet} for the 2020s.
\newblock In \emph{Proceedings of the IEEE/CVF Conference on Computer Vision
  and Pattern Recognition}, pp.\  11976--11986, 2022.

\bibitem[Lucic et~al.(2018)Lucic, Kurach, Michalski, Gelly, and
  Bousquet]{lucic2018gans}
Lucic, M., Kurach, K., Michalski, M., Gelly, S., and Bousquet, O.
\newblock Are {GAN}s created equal? {A} large-scale study.
\newblock In \emph{Advances in Neural Information Processing Systems},
  volume~31, 2018.

\bibitem[Owen(2013)]{owen2013monte}
Owen, A.~B.
\newblock \emph{Monte Carlo theory, methods and examples}.
\newblock Self-published, 2013.

\bibitem[Picard(2021)]{picard2021torch}
Picard, D.
\newblock Torch.manual\_seed(3407) is all you need: On the influence of random
  seeds in deep learning architectures for computer vision.
\newblock \emph{arXiv preprint arXiv:2109.08203}, 2021.

\bibitem[Poyarkov et~al.(2016)Poyarkov, Drutsa, Khalyavin, Gusev, and
  Serdyukov]{poyarkov2016boosted}
Poyarkov, A., Drutsa, A., Khalyavin, A., Gusev, G., and Serdyukov, P.
\newblock Boosted decision tree regression adjustment for variance reduction in
  online controlled experiments.
\newblock In \emph{Proceedings of the 22nd ACM SIGKDD International Conference
  on Knowledge Discovery and Data Mining}, pp.\  235--244, 2016.

\bibitem[Rosenbaum(1984)]{rosenbaum1984consequences}
Rosenbaum, P.~R.
\newblock The consequences of adjustment for a concomitant variable that has
  been affected by the treatment.
\newblock \emph{Journal of the Royal Statistical Society: Series A},
  147\penalty0 (5):\penalty0 656--666, 1984.

\bibitem[Sellam et~al.(2022)Sellam, Yadlowsky, Tenney, Wei, Saphra, D'Amour,
  Linzen, Bastings, Turc, Eisenstein, Das, and Pavlick]{sellam2022multiberts}
Sellam, T., Yadlowsky, S., Tenney, I., Wei, J., Saphra, N., D'Amour, A.,
  Linzen, T., Bastings, J., Turc, I.~R., Eisenstein, J., Das, D., and Pavlick,
  E.
\newblock {MultiBERTs}: {BERT} reproductions for robustness analysis.
\newblock In \emph{International Conference on Learning Representations}, 2022.

\bibitem[Touvron et~al.(2021)Touvron, Cord, Douze, Massa, Sablayrolles, and
  J{\'e}gou]{touvron2021deit}
Touvron, H., Cord, M., Douze, M., Massa, F., Sablayrolles, A., and J{\'e}gou,
  H.
\newblock Training data-efficient image transformers \& distillation through
  attention.
\newblock In \emph{International Conference on Machine Learning}, pp.\
  10347--10357. PMLR, 2021.

\end{thebibliography}
\bibliographystyle{icml2026}

\appendix

\section{Raw Run Variability}
\label{app:variability}

\begin{table}[h]
\caption{Mean test accuracy (\%) and run-to-run standard deviation (pp) across
50 runs per cell.}
\label{tab:variability}
\vskip 0.1in
\begin{center}
\begin{small}
\begin{tabular}{@{}lrr@{}}
\toprule
\textbf{Configuration} & \textbf{Mean acc.} & \textbf{Std (pp)} \\
\midrule
CIFAR-10 / ResNet-18 & 94.68 & 0.14 \\
CIFAR-10 / ViT-Tiny & 93.68 & 0.23 \\
CIFAR-10 / ConvNeXt-Tiny & 96.61 & 0.13 \\
\midrule
CIFAR-100 / ResNet-18 & 77.04 & 0.24 \\
CIFAR-100 / ViT-Tiny & 70.06 & 0.46 \\
CIFAR-100 / ConvNeXt-Tiny & 79.34 & 0.33 \\
\midrule
Tiny-ImageNet / ResNet-18 & 65.42 & 0.27 \\
Tiny-ImageNet / ViT-Tiny & 55.55 & 0.44 \\
Tiny-ImageNet / ConvNeXt-Tiny & 67.82 & 0.69 \\
\bottomrule
\end{tabular}
\end{small}
\end{center}
\end{table}

\end{document}